\documentclass{sigchi-latex-proceedings/sigchi}

 \toappear{
 Permission to make digital or hard copies of all or part of this work
 for personal or classroom use is granted without fee provided that
 copies are not made or distributed for profit or commercial advantage
 and that copies bear this notice and the full citation on the first
 page. Copyrights for components of this work owned by others than ACM
 must be honored. Abstracting with credit is permitted. To copy
 otherwise, or republish, to post on servers or to redistribute to
 lists, requires prior specific permission and/or a fee. Request
 permissions from \href{mailto:Permissions@acm.org}{permissions@acm.org}. \\
 \emph{CHI PLAY '18,  October 28--31, 2018, Melbourne, VIC, Australia} \\
 \copyright~2018 Association of Computing Machinery. \\
 ACM ISBN 978-1-4503-5624-4/18/10\ ...\$15.00. \\
 DOI: \url{https://doi.org/10.1145/3242671.3242700}
 }

\usepackage{balance}       % to better equalize the last page
\usepackage{graphics}      % for EPS, load graphicx instead 
\usepackage[T1]{fontenc}   % for umlauts and other diaeresis
\usepackage{txfonts}
\usepackage{mathptmx}
\usepackage[pdflang={en-US},pdftex]{hyperref}
\usepackage{color}
\usepackage{booktabs}
\usepackage{textcomp}
\usepackage{multirow}

\usepackage{microtype}        % Improved Tracking and Kerning
\usepackage{ccicons}          % Cite your images correctly!
\usepackage{todonotes}

\usepackage{tabularx}
\usepackage{enumitem}
\usepackage{comment}
\def\plaintitle{Informing a BDI Player Model for an Interactive Narrative}
\def\plainauthor{Jessica Rivera-Villicana, Fabio Zambetta, James Harland,
  Marsha Berry}

\def\plainkeywords{Player Modelling; BDI; User Modelling; Interactive Narratives; Player Profiling;}

\makeatletter
\def\url@leostyle{%
  \@ifundefined{selectfont}{
    \def\UrlFont{\sf}
  }{
    \def\UrlFont{\small\bf\ttfamily}
  }}
\makeatother
\def\pprw{8.5in}
\def\pprh{11in}

\definecolor{linkColor}{RGB}{6,125,233}
\hypersetup{%
  pdftitle={\plaintitle},
  pdfauthor={\plainauthor},
  pdfkeywords={\plainkeywords},
  pdfdisplaydoctitle=true, % For Accessibility
  bookmarksnumbered,
  pdfstartview={FitH},
  colorlinks,
  citecolor=black,
  filecolor=black,
  linkcolor=black,
  urlcolor=linkColor,
  breaklinks=true,
  hypertexnames=false
}

\begin{document}

\title{\plaintitle}

\numberofauthors{1}
\author{%
  \alignauthor{Jessica Rivera-Villicana\textsuperscript{1}, Fabio Zambetta\textsuperscript{1}, James Harland\textsuperscript{1}, Marsha Berry\textsuperscript{2}\\
  \affaddr{\textsuperscript{1} School of Science, \textsuperscript{2} School of Media and Communication}\\
  \affaddr{RMIT University}\\
  \affaddr{Melbourne, Australia}\\
  \email{\{jessica.riveravillicana, fabio.zambetta, james.harland, marsha.berry\}@rmit.edu.au}}\\ 
  %\emptyauthor
  %\alignauthor{Leave Authors Anonymous\\
  %  \affaddr{for Submission}\\
  %  \affaddr{City, Country}\\
  %  \email{e-mail address}}\\
  %\alignauthor{Leave Authors Anonymous\\
  %  \affaddr{for Submission}\\
  %  \affaddr{City, Country}\\
  %  \email{e-mail address}}\\
  %\alignauthor{Leave Authors Anonymous\\
  %  \affaddr{for Submission}\\
  %  \affaddr{City, Country}\\
  %  \email{e-mail address}}\\
}

\maketitle

\begin{abstract}
   This work focuses on studying players behaviour in interactive narratives with the aim to simulate their choices. Besides sub-optimal player behaviour due to limited knowledge about the environment, the difference in each player's style and preferences represents a challenge when trying to make an intelligent system mimic their actions. Based on observations from players interactions with an extract from the interactive fiction Anchorhead, we created a player profile to guide the behaviour of a generic player model based on the BDI (Belief-Desire-Intention) model of agency. We evaluated our approach using qualitative and quantitative methods and found that the player profile can improve the performance of the BDI player model. However, we found that players self-assessment did not yield accurate data to populate their player profile under our current approach.
\end{abstract}

\graphicspath{{Imgs/}}




%Copy-paste your ACM 2012 classifiers above. See instructions: \url{https://www.acm.org/publications/class-2012}.  This section is required.


\section{Introduction}
\label{sec:introduction}
In this paper, we present an approach to player modelling for Interactive Narratives (INs), focusing on the task of simulating the behaviour of different players. Achieving this requires a thorough analysis of players motivation and gaming style.
Commercial Role-Playing Games (RPGs) such as The Witcher series~\cite{CDProjectRED} or Heavy Rain~\cite{Dream2010} have shown that implementing an IN offers benefits such as freedom of choice, and the sense of these choices affecting the outcome of the story~\cite{riedl2013interactive}.  
As opposed to linear narratives, INs provide different versions of a story depending on the choices made by each player. %While this is a desirable feature, it also represents a challenge when trying to predict the behaviour of a specific human player.
Game creators seek information about players in order to increase or maintain players interest. This can be achieved in different ways, such as adjusting the level of difficulty to match the players skill level~\cite{booth2009ai}, more believable Non-Player Characters (NPCs)\cite{Karpov2012}, suggesting micro-transactions~\cite{6374152} or content personalisation~\cite{Yannakakis2011}. The study of players interaction with games is known as player modelling, and it implements knowledge from areas such as psychology, artificial intelligence, human-computer interaction and user modelling.

As the story becomes richer, the number of potential story versions in an IN grows exponentially, making them intractable for a human. Consequently, it becomes infeasible for authors to ensure all the resulting stories meet their quality standards~\cite{riedl2013interactive}. 
From the author's perspective, player modelling can be beneficial for the creation of tools to assist them with the design and validation of the quality of the narrative~\cite{Bida2013,Liapis2013}. A player model with accurate representations of real players goals and behaviour can provide authors with simulations of players to make sure that their story is diverse enough to cater to all preferences. Simulated player behaviour can also help creating archetypes for characters and NPCs within the IN (e.g. the behaviour of a thief in situations not considered by the author).
From a player's perspective, player models can help personalising the content of an IN. Knowing what the current player will do can help an intelligent narrative system make decisions to make sure they discover the story considered best for them, while preserving the narrative quality intended by the author~\cite{sharma2010drama}.
The implementation of authoring systems or drama managers is outside of the scope of this paper. However, our work in understanding and modelling player behaviour can help bridge the gap between the state of the art and such systems.

 There are several challenges involved in simulating the behaviour of a specific player in an IN, one of them being sub-optimal player behaviour as a result of their partial awareness of the state of the environment, the main goal of the game, and the outcome of their actions~\cite{Rivera-Villicana2016}. In addition, we face the challenge of reproducing the gaming style of different players. This requires a model that encodes factors that define their behaviour, which we assume is driven by individual preferences combined with their interpretation of the game's goals. Our empirical study shows that players make different assumptions depending on their experience with games with INs, this is discussed in more detail in our findings section. %~\ref{sec:findings}.

We adopt a Belief-Desire-Intention (BDI) model to emulate human behaviour in Anchorhead, an interactive fiction used in IN research~\cite{Rivera-Villicana2016, sharma2010drama, Nelson2006}. Since BDI is based on folk psychology (the way we think others think), we believe that implementing a player profile (a static set of features describing the player) can help us differentiate the decisions made by players with different styles while emulating players sub-optimal behaviour in this game genre. 
Our specific research aims in this work are:
\begin{enumerate}[noitemsep]
  \item{To identify elements that define individual players behaviour in an IN game to build a player profile.}
  \item{To evaluate the performance of a BDI model informed with such player profile compared to an uninformed BDI model.} 
\end{enumerate}

While the player model developed is applied to a specific narrative, we believe that our methodology can be generalised to create player models for other INs using the same player profile to inform them, as the mechanics of Anchorhead are present in most INs. Our contributions include:
\begin{enumerate}[noitemsep] 
  \item{A player profile with four elements that can inform the decision process of a BDI agent to emulate the behaviour of different players.}
  \item{A methodology to create a BDI player model for an IN using a player profile to simulate the behaviour of players with different styles and preferences.}
  \item{Insights on players behaviours and strategies as an outcome of our analysis of players behaviour via semi-structured interviews.}
  \item{An analysis of the reliability of players to quantify their behaviour and preferences via questionnaires.}  
\end{enumerate}

To the best of our knowledge, our  study aiming to simulate players behaviour in the IN domain is one of the first attempts towards providing better player experiences in the genre of INs. 
The remainder of this paper is organised as follows: we introduce the concepts and terminology related to our work in the background section. We then explain our methodology to create a BDI player model and inform it with the proposed player profile. In the analysis section, we explain how we evaluated the accuracy of our model and our findings. We then present our conclusions and propose future work based on our observations.

\begin{figure*}
\centering
\includegraphics[width=0.8\linewidth]{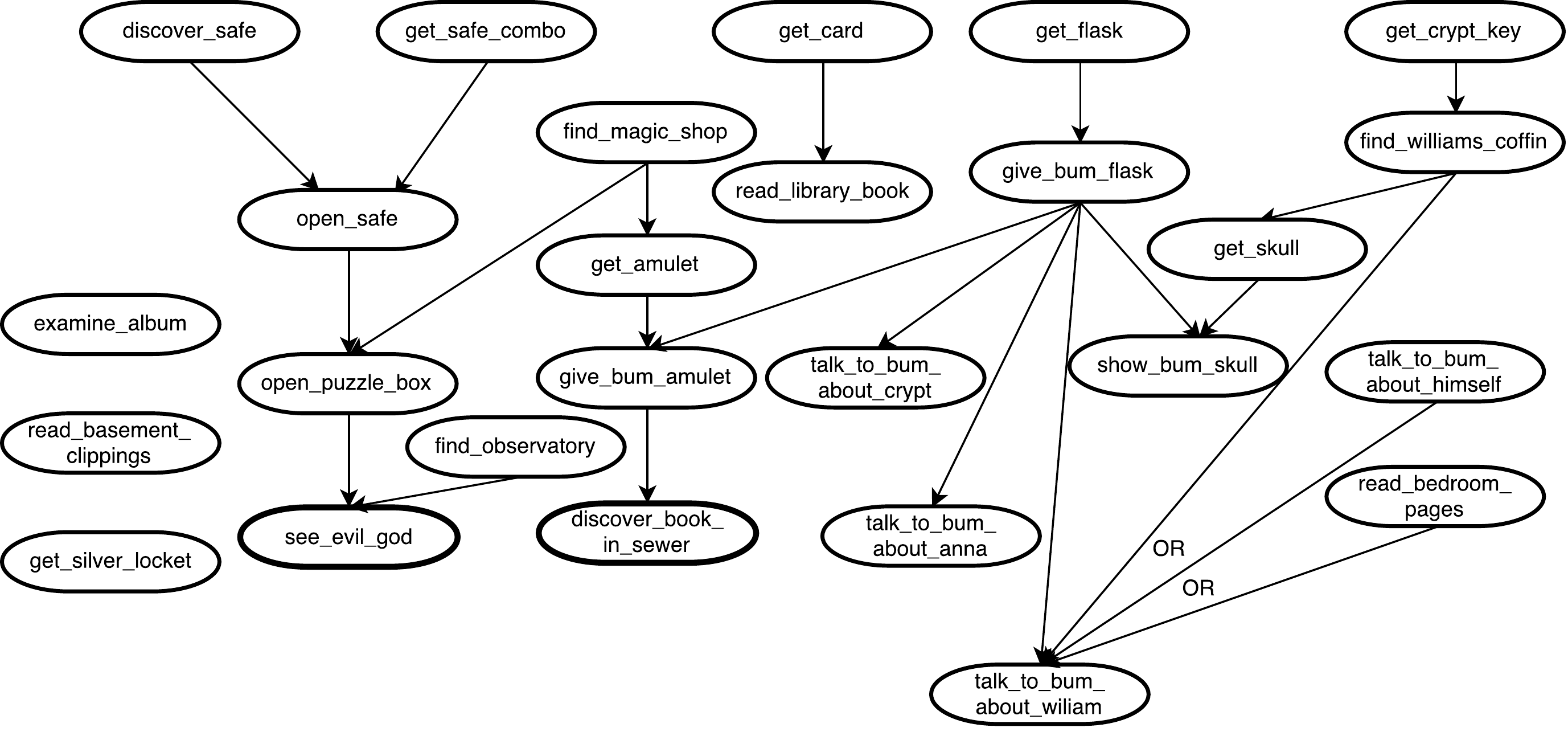} 
	\caption{Plot graph of Anchorhead's extract used in this work.}
	\label{fig:plotgraph}
\end{figure*}

\section{Background}
\label{sec:background}

In this section we introduce the concepts of interactive narrative, player modelling and the BDI model. We also describe Anchorhead and discuss previous work related to our research.

\subsection{Interactive Narratives}
An Interactive Narrative (IN) is a form of storytelling where the listener (reader, player or user) is given freedom to affect the direction of the story by choosing how to act at different stages~\cite{riedl2013interactive}. In some cases, the choices made only affect the order of some events (e.g. the order in which a player asks questions to a Non-Player-Character). However, some of these choices have immediate or delayed consequences (e.g., choosing to join the enemy can cause our current friends to be hostile insantly, or not asking a specific question can cause a mission to fail after some time because the answer was crucial to finish it).
A modern example of the implementation of an IN is the game \textit{The Witcher 2}~\cite{CDProjectRED}. However, INs have been around for a long time in different formats, such as the ``choose your own adventure'' books, or interactive fictions (text-based adventure games)~\cite{lebowitz2011interactive}.

To provide freedom of choice, INs are structured using~\textit{plot points}, i.e., important events in the story that may or may not depend on other events. The author of the IN can establish precedence constraints betweeen plot points without confining the player to a specific path. For instance, opening a locked door has the constraint that the key must be found and collected first, but the player can A) collect the key, talk to a character, start and complete a new mission in a different location, come back and unlock the door, or B) collect the key and immediately unlock the door. One common way to represent precedence constraints is a \textit{plot graph} (see figure~\ref{fig:plotgraph}); a directed, acyclic graph, where the nodes represent important plot points, and the arcs represent order constraints defined by the author to ensure the quality of the story discovered by the player~\cite{weyhrauch1997guiding, doi:10.1162/pres.1993.2.1.1}. 

As the options available to the player grow, the number of possible paths between collecting the key and unlocking the door in the previous example becomes larger. If we consider the possible paths between the beginning and the end of the game, picking one that exactly matches the path discovered by a specific player becomes a challenging task. This comes down to making the same decisions as the player in question. A player model can aid this decision-making process, provided it encodes information representative of the player and makes good use of it. Player modelling is the use of computational intelligence to build models of players interaction with games~\cite{Yannakakis2013}. Player models can also include player profiles, a collection of static information not necessarily related to the game, such as cultural background, gender, age, etc.

\subsection{Anchorhead}
\label{sec:anchorhead}
Anchorhead is an interactive fiction written by Michael S. Gentry in 1998~\cite{anchorhead}. The story takes place in the fictional town of Anchorhead, where the protagonist and her husband inherited a mansion from distant family. The previous owner of the house, Edward Verlac, killed his family and committed suicide. From this point, the player needs to explore the world by visiting places, examining objects, talking to characters, etc. to find out the truth behind the Verlac family. 
The original story is divided into five days, with over one hundred relevant plot-points \cite{Nelson2005}. In this project, we use only the second day, which contains enough constraints and plot points to make a sub-story that has been used in player modelling research before \cite{sharma2010drama}.

In this extract, the player can reach two possible endings, identified with bold frames in figure \ref{fig:plotgraph}.
The order in which the plot points are revealed depends on the players actions as well as the current game context. During a game, players will discover plot points from both sub-stories as they explore the world, consequently, the final story discovered by each player in each run of the game is expected to be different.

\subsection{The BDI Model of Agency}
\label{sec:bdi}
The Belief-Desire-Intention (BDI) model is a framework commonly used to build intelligent agent systems~\cite{padgham2005developing} based on the model of practical reasoning developed by Bratman~\cite{bratman1987intention}. The philosophy behind this model is that humans interact with their environment using three main characteristics: A set of \textit{beliefs} composed by the information sensed about the environment, a set of \textit{desires} that constitute the motivational aspect of reasoning (i.e., the goals that will drive our behaviour) and a set of \textit{intentions}, a selection of plans to achieve our current goals given our current beliefs.
Considering that an IN requires the player to make a choice given the conditions known from the environment, we can model IN players as BDI agents using the state of the game as beliefs (which change as the player advances throughout the story), pre-defined goals depending on such beliefs, and a library of plans for achieving such goals as intentions.

To design the player model, we followed the Prometheus methodology~\cite{Winikoff2004}, an approach commonly used to design BDI agents. Prometheus suggests decomposing the agent's intended behaviour into goals and sub-goals, represented as a \textit{goal diagram}, or goal tree diagram (see figure~\ref{fig:ourgoaldiagram}). Note that a goal diagram and a plot graph represent different concepts.
Implementations of BDI allow to diversify behaviour by using \textit{context conditions} for plans, i.e, statements to verify when a plan can be executed depending on the current beliefs. Context conditions are useful to define different plans to achieve the same goal. For instance, if the goal is to move to a specific location, we can take the shortest path or explore the world until we find the intended place. Which plan we execute depends on whether we know how to get to the destination, whether we have already explored the locations between the source and target locations and whether we like exploring the environment.

\subsection{Related work}
\label{sec:relatedwork}
We now introduce an overview of related work that tackles the player modelling problem from different perspectives.

PaSSAGE (Player-Specific Stories via Automatically Generated Events) is a drama manager that categorises players in order to select the event that is most suitable for the category of the current player ~\cite{Thue2007}. Another work that uses information from players to in order to make decisions regarding drama management is that in~\cite{yu2012sequential}, which uses a recommendation system based on player's feedback, under the assumption that players who share preferences in the past are likely to share them in the future. One more approach of using a player model to inform drama management is the use of case based reasoning and players' feedback to validate the suggestions made by a drama manager~\cite{sharma2010drama}.

The work in~\cite{wang2017simulating} shows an approach to simulate players behaviour using a long short-term memory neural network. This work explains the benefits of using simulated players in the domain of IN planners (a form of drama manager). Such benefits include reducing the dependency on large amounts of real players observations for training and exploiting the logics of the IN to access states not commonly visited by human players. The work in~\cite{wang2017simulating} uses similar player descriptors to our approach, such as gaming experience and context knowledge to generate their variety of players behaviour.

Even though these works make use of a player model, their focus is on drama management. Our work in contrast focuses on trying to mimic the behaviour of different players without performing drama management. Although we reiterate that our work can be applied as a way to predict player choices to support a drama manager, our motivation is not limited to that application.

BDI has been used in video games research to define different forms of intelligent systems, such as a game master that creates a narrative experience from modular events~\cite{luong2013bdi}. The difference between the work in~\cite{luong2013bdi} and ours is where the BDI model is applied; in their work, the BDI agent is the game master that makes decisions and it does not perform player modelling. In our work, the BDI agents are implementations of our player model that are expected to behave like real players.

The work in~\cite{Peinado2008} presents a BDI model that considers the pre-defined personality of Non-Player Characters (NPCs) to define their reactions to changes in the environment, as well as the execution of certain plans. Although there is certain similarity with our work, the intent in~\cite{Peinado2008} is to create appropriate behaviour for NPCs rather than emulating the behaviour of real players.

Another BDI application in games is the player model for a first-person shooter in~\cite{Norling2003}. Although this work is not implemented on an IN domain, it provides insights on how the BDI paradigm can be used to model players in virtual worlds. In~\cite{Norling2004a}, a BDI approach is used to create interactive NPCs with realistic behaviour. Although that work is not implemented in INs either, it shows how we can model BDI agents that resemble humans in video games.

In our previous work, we used the BDI model to imitate human behaviour in INs~\cite{Rivera-Villicana2016}. We shown that the model proposed can produce behaviour that resembles human choices better than direct pathways to either of the ends, under the assumption that humans do not behave optimally in INs. In this work, our aim is to investigate if we can increase the ability of that model to emulate the behaviour of different players by using measurements of their play style and preferences.

\section {BDI Player Model for Anchorhead}
\label{sec:methodology}

In previous work, we explained how players behaviour in INs tends to be sub-optimal, and how creating a BDI Player Model (BDI PM) for this genre requires modelling a set of scattered goals that get triggered as the player discovers information, rather than a goal tree that connects most sub-goals in order to achieve a main goal (usually unknown by IN players)~\cite{Rivera-Villicana2016}. The fact that many of these goals are common among most INs allows us to define a baseline to model players behaviour before modelling goals specific for each story. Mechanics such as exploring rooms, finding and taking objects to use them later, as well as interacting with characters to learn information are typical in this genre.

\begin{figure*}
\centering
\includegraphics[width=\linewidth]{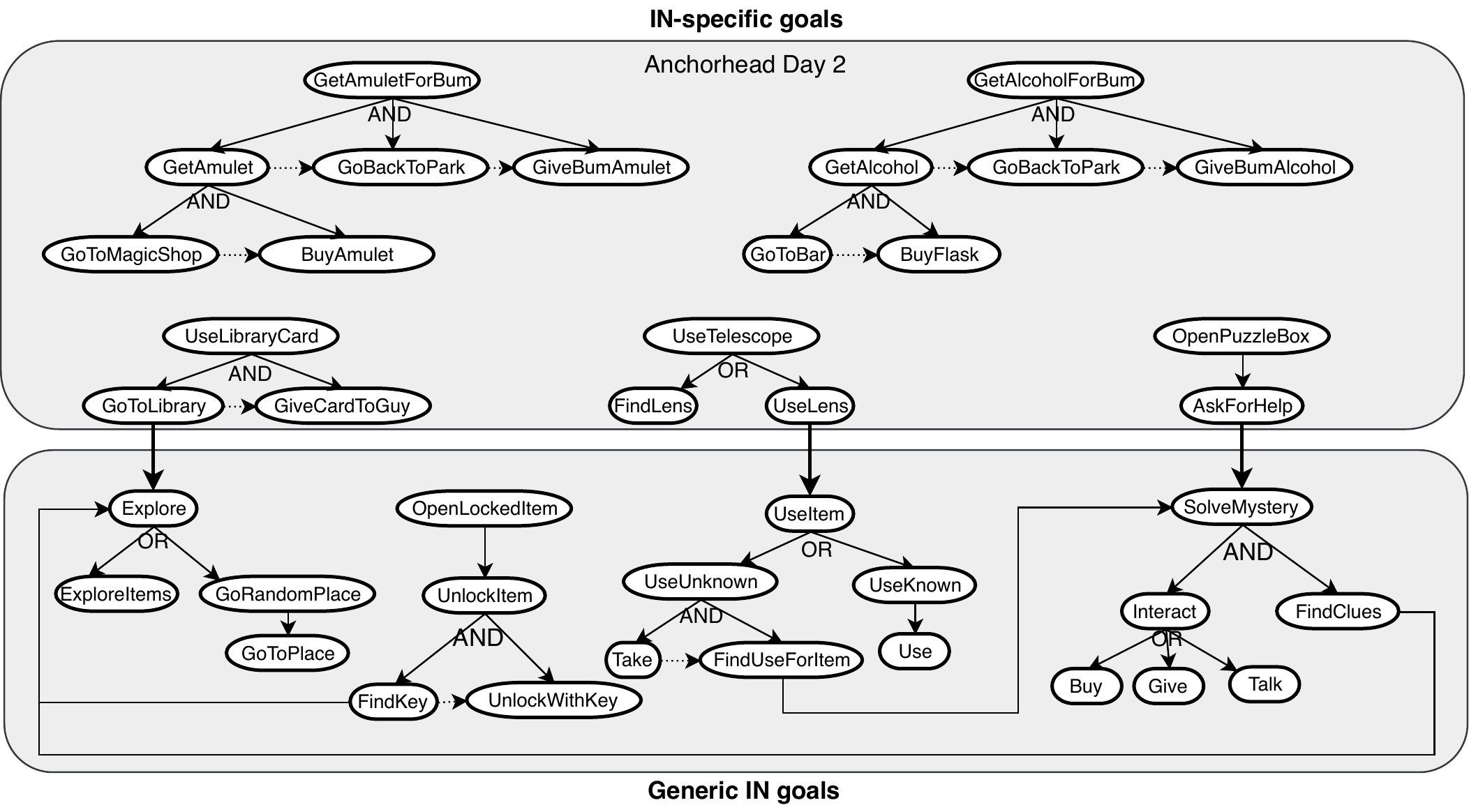} 
	\caption{Goal diagram used for Anchorhead.}
	\label{fig:ourgoaldiagram}
\end{figure*}

By following the Prometheus methodology~\cite{padgham2005developing}, we can define a BDI agent to simulate players with goals and plans that keep triggering new goals until the game is eventually finished. In this case, the goals triggered at the beginning of the game are ``low-level'', such as exploring the current location or talking to a character. As the player approaches the end of the game, ``higher-level'' goals are discovered that use the former (e.g., defeat a NPC, who just escaped after finding out he is the enemy, triggering the sub-goal of finding the NPC first, which depends again on navigation, interaction, etc.). 
We can then define a generic ``low-level'' player behaviour and augment it with the ``higher-level'' goals specific to each IN. Figure~\ref{fig:ourgoaldiagram} shows the goal diagram design of a BDI player model for the version of Anchorhead used in our work and how we divided its goals into generic IN (or low-level) goals and IN-specific (or higher-level) goals.

\subsection{Player profile}
\label{sec:playerprofile}

Our previous work also reported that the behaviour of a BDI agent in an IN is mainly defined by four elements: goal priorities, navigation, action selection when the agent is idle and decisions made regarding the objects found~\cite{Rivera-Villicana2016}. This led us to define a Player Profile (PP) that can be used by the BDI PM to refine the behaviour according to each player's personality.
We analysed the 22 traces collected in that work to identify elements that we can use in the BDI PM to generate more player-specific game traces. We took the observations in~\cite{sharma2010drama} regarding players experience and previoius knowledge of the game as a starting point to identify the kind of player in each trace. We also considered the player categories used in PaSSAGE~\cite{Thue2007}.

We considered ratios between types of actions, length of traces and plot points to identify patterns. For instance, to determine whether a trace belonged to an explorer player type, we considered the number of times the action `examine' was issued versus the total number of actions in the trace. Although this approach did not yield a normalised measurement, it was helpful to hypothesise how different factors of players personalities are reflected in their traces. We found that it was difficult to tag a trace with a specific style because the behaviour tends to be different depending on the current goal of a player. The four characteristics that we identified as low or high in most traces are:

\begin{itemize}
\item{\textbf{Familiarity with the game:} When a player is already familiar with a game of this nature, we would not expect them to perform exactly the same actions when re-playing it; the knowledge acquired in previous games should affect their decisions and result in a substantially different trace. This factor could help us in the process of selecting the next action whenever the agent is idle, as we can presume that if there exists a trace file for that player, the current game will present different actions than the existing one in order to discover different plot-points, and potentially a different ending.}
\item{\textbf {Gaming experience:} As previous research suggests, experienced gamers tend to identify important objects or information and use strategies more often than novices  because of their previous experience \cite{sharma2010drama}. By considering gaming experience in our model, we can define in a better way the decision making about the objects in the game world. }
\item{\textbf{Preference to explore:} This element can help us refine the way the agent moves inside the IN world. If we want to mimic the behaviour of someone with a high preference to explore, a navigation algorithm that exhausts the possible places to go during a game would be the right choice. However, if the player prefers not to explore and go straight to solve quests, then we should use a more direct navigation method between two places in order to accurately reproduce that behaviour.}
\item{\textbf{Persistence:} Some players give up on their intentions earlier than others. The reasoning behind this decision also varies depending on how each player interprets the state of the game. Sometimes a new goal seems more important than the ones being pursued, other times they consider they have spent more time than they expected attempting one goal, or a new element in the environment catches their attention and they prefer to continue exploring rather than focusing on their current goals. This factor helps us have a better prioritisation of an agent's goals. For instance, if the current player's persistence is low, then the agent could give more priority to explore than to solve puzzles, or work on the goals acquired recently and drop the ones discovered earlier in the game.}
\end{itemize}

To design a BDI PM for any other IN, we propose a methodology that bridges the assumption that IN players have a sub-optimal behaviour and the guidelines in the Prometheus methodology to design an agent whose behaviour is preferably optimal. Our approach suggests the use of the player profile to achieve a diversified behaviour that in many cases will not be optimal. Each of the stages in our approach aligns with one stage of Prometheus~\cite{padgham2005developing}, indicated between brackets in the following step-by-step methodology.

\begin{enumerate}[label=\textbf{Stage \Roman*:}, align=left]
\item{\textbf{Identification of requirements (System specification)}}
\begin{enumerate}[label=(\Roman{enumi}.\arabic*)]
\item{Identify basic actions to interact with the game that are not covered by the generic IN goals from figure~\ref{fig:ourgoaldiagram}.}
\item{Identify the plot points that require a special set of actions and information specific to  this IN.\\
Example: A box cannot be opened with a key and requires the player to ask an NPC for help.}
\item  \label{example:locksmith}{Identify how a player would set goals from such plot points with incomplete information.\\
Example: If the player realises the box cannot be opened with keys and sees an axe in the room, a goal `openBoxWithAxe' might be defined as a sub-goal of `openBox'. The player does not know there is a locksmith in a different room. If the player has already visited the locksmith, they will most likely go to him, therefore, an alternative goal `takeBoxToLocksmith' is also needed.}
\item{Expand the generic IN goal diagram with the actions and goals identified (IN-specific goals in figure~\ref{fig:ourgoaldiagram}).}
\end{enumerate}

\item{\textbf{Coverage verification (System specification)}}
\begin{enumerate}[label=(\Roman{enumi}.\arabic*)]
\item{Verify that all the plot points in the plot graph can be reached by the goals defined in the goal diagram.}
\item{Identify the information that needs to be stored as beliefs and how it is modified.\\
Example: A door is visible, and adding this to the belief set will trigger a goal `openDoor'.}
\end{enumerate}

\item{\textbf{Implementation (Detailed design)}}
  \begin{enumerate}[label=(\Roman{enumi}.\arabic*)]
  \item{Identify possible relationships between sub-goals and factors of the player profile.\\
  Example: A player with low persistence may be more likely to reach the goal `openBoxWithAxe' even if they know about the locksmith from the example in step~\ref{example:locksmith}. However, a persistent player is more likely to take the extra steps involved in `takeBoxToLocksmith'.}
\item{Implement plans for each goal and sub-goal.}
\end{enumerate}
\end{enumerate}

\begin{figure*}
\centering
\includegraphics[width=\linewidth]{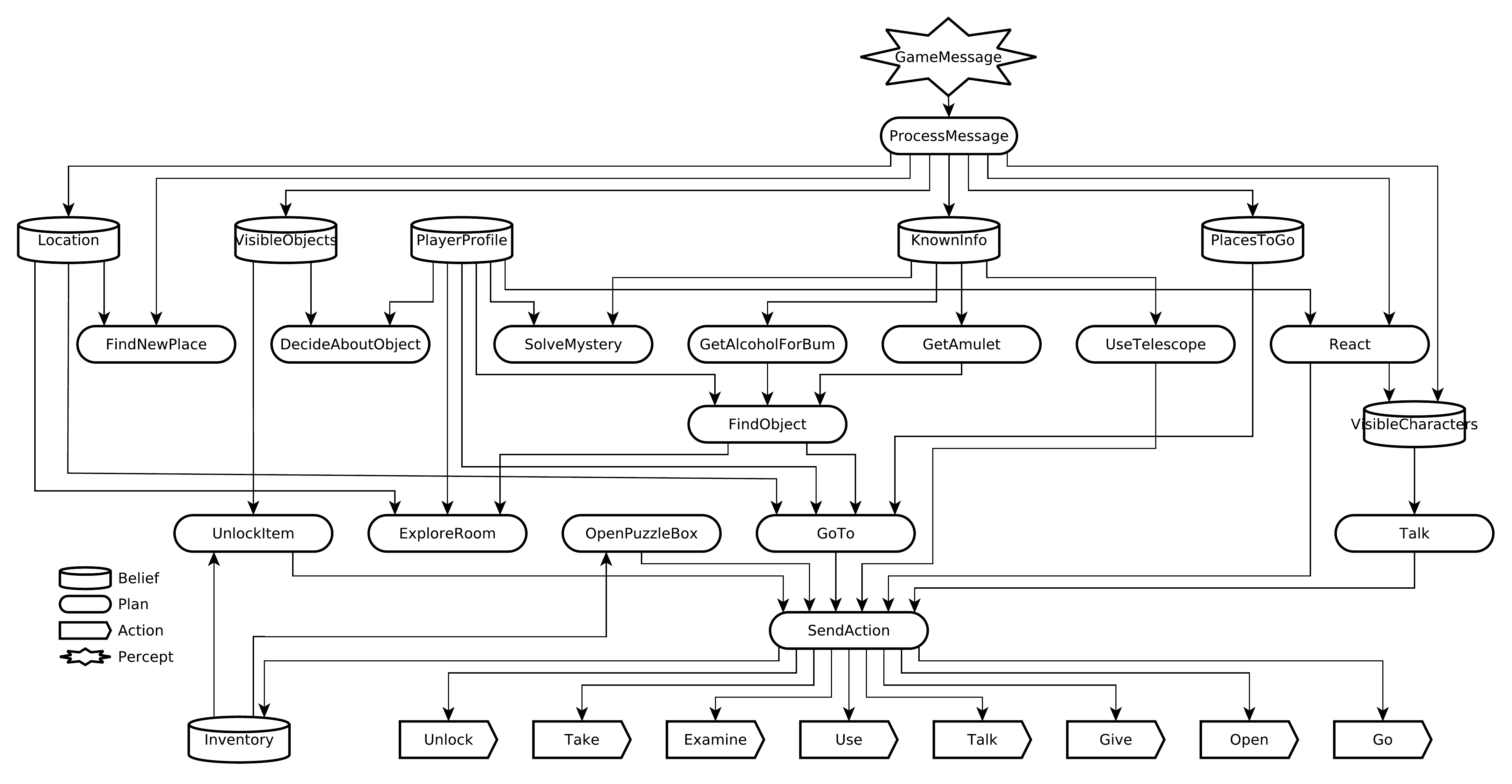} 
	\caption{Overview diagram of our informed player model.}
	\label{fig:diagramOverview}
\end{figure*}

\subsection{Informing the BDI player model to refine plan selection}

We implemented the factors in the PP as part of the belief set of the generic BDI PM, whose instance we call Uninformed Agent (UA), to be used as plan context conditions. For the goals related to the factors that define the agent's behaviour, explained in the previous section, we now have at least two plans that depend on the PP, one for a low score and one for a high score of the factor in question. We refer to the BDI PM that uses the PP as informed BDI PM, and to its instance as Informed Agent (IA).
Figure \ref{fig:diagramOverview} shows the overview diagram of the BDI informed PM designed for Anchorhead. As the diagram indicates, plans used to define behaviour such as ``DecideAboutObject'' or ``ExploreRoom'' take into account the values of the player profile, which are filled for each player before their simulation. 
We use the threshold of 0.5 to decide which plan to execute depending on the normalised factor in the PP for two reasons:
\begin{itemize} [noitemsep]
\item{Our approach to collect players information, described in our experimental setup, is designed to give normalised values higher than 0.5 if the PP elements are high enough as considered by the authors.}
\item{To reduce the possibility of dealing with contradicting plans. Having four elements in the PP to define when to trigger a plan allows for evaluating more than one factor as context conditions. While this facilitates a more detailed PM, the risk of launching two plans that contradict each other increases (e.g. go to a different room to find a key because a chest was found and the player is persistent and keep exploring the current room because the player is also an explorer). Having only two possibilities per element ($low:{0 \leq n \leq 0.5}$ and $high:{0.5<n \leq 1}$) helps keeping the number of plans manageable.}
\end{itemize}

\begin{figure}[t]
\centering
\includegraphics[width=\linewidth]{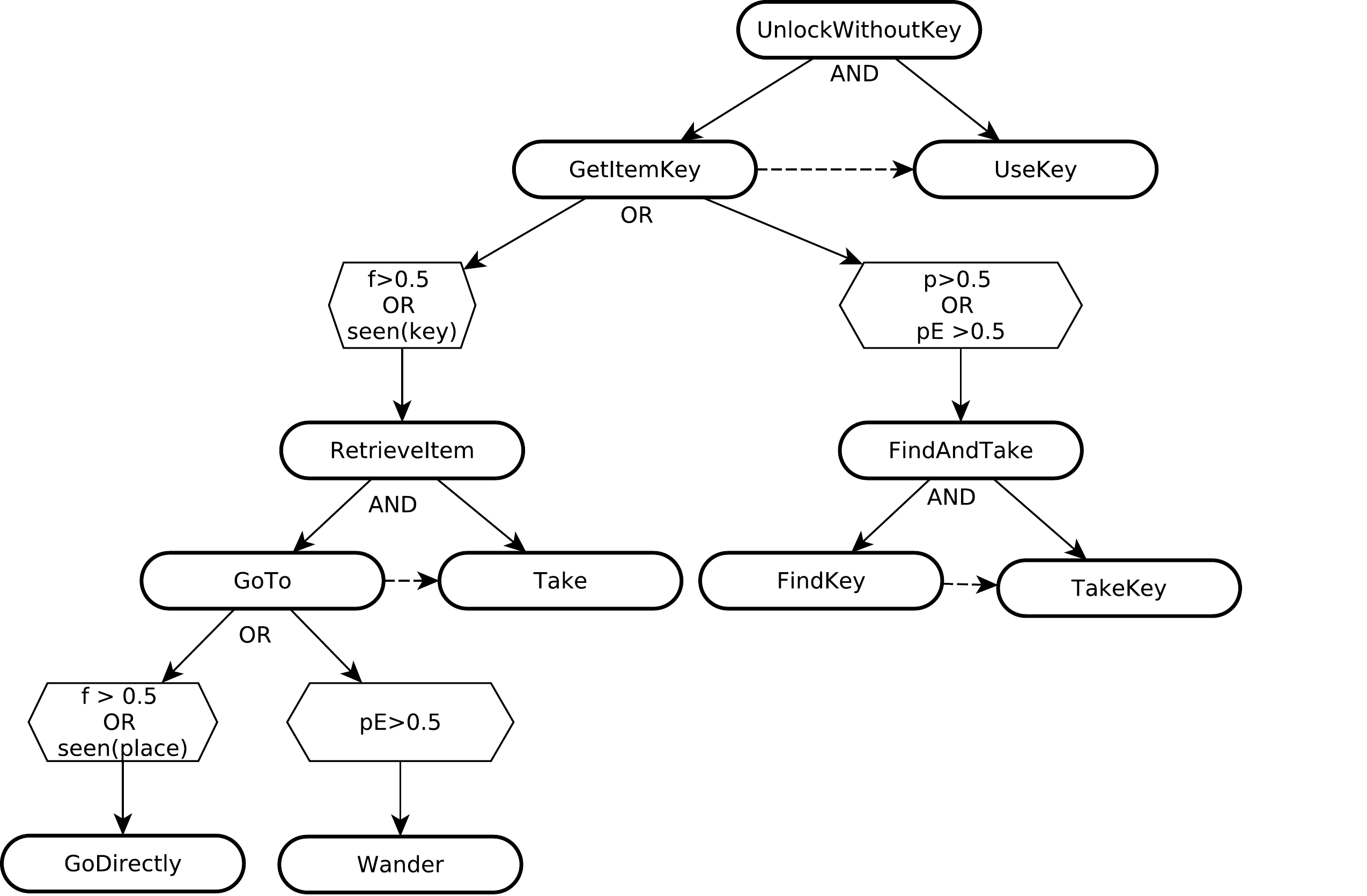} 
\caption{Example of the use of context conditions from the player profile.}
\label{fig:contextConditions}
\end{figure}

In the case of persistence, we considered the approximate length of a simulated game (50-60 seconds) to decide for how long a plan should be attempted before dropping it. The time limit we chose was ten seconds. 
At a low level, specific plans are triggered depending on the values of the PP. Figure \ref{fig:contextConditions} is an example of the steps to find the key for an item, depending on the factors of the PP and the current game context.

\section{Analysis}
\label{sec:evaluation}

We used three elements for our analysis: game traces collected from players, information obtained from semi-structured interviews, and performance measurements of the BDI agents described in the previous section. %Section~\ref{sec:methodology}.
The process we followed and our findings are reported in this section.

\subsection{Semi-structured interviews}
\label{sec:interviews}
Up to this point, the PP is only based on observations from quantitative data (actions during game play). We consider the need to validate the PP design with an insight on the players motivation while playing. At the same time, we want to make sure that there are no factors in players behaviour which can be relevant for a BDI model being left out in our method.
We consider semi-structured interviews an appropriate artifact to gain qualitative evidence to make these validations. Previous work has followed a similar approach to evaluate a player model that is part of a drama management system~\cite{sharma2010drama}. The difference between that work and ours is that we are aiming to identify patterns of behaviour, instead of evaluating quality of experience.
The following scenario helps us illustrate our need of a more detailed explanation of a player's strategies. Given the following trace:

\textit{\small{[examine-album, get-card, get-safe-combo, open-safe, get-crypt-key, get-silver-locket, read-basement-clippings, read-bedroom-pages, find-williams-coffin, see-skull, get-skull, start-talking-to-bum, get-flask, give-bum-flask, ask-bum-about-photo, get-library-book, no-more-flasks, find-magic-shop, buy-magic-ball, get-amulet, ask-bum-about-william, open-puzzle-box, show-bum-skull, give-bum-amulet, reject-amulet-from-player, discover-book-in-sewer]}}\\

We have several theories to classify this player's behaviour:\\

\begin{itemize}[noitemsep]
\item{The player has a high preference to explore: They revealed a large number of plot-points compared to the rest of the players in the pilot experiment.}
\item{The player was looking for a specific item or place: If the player could not find such item or location, they might have unintentionally discovered all these plot points, but this does not necessarily mean the player has a high preference to explore, but rather a high persistence.}
\item{The player was not following any specific strategy and tried random actions until they eventually finished the game.}
\end {itemize}

	Our player profile covers these possibilities; preference to explore covers the first scenario, persistence can be translated into constantly looking for an item to achieve a goal and perhaps a low gaming experience results in a lack of strategy to finish the game. However, it is unlikely that only one of the above scenarios is true while the rest are false. Judging from observation and gaming experience, we believe that the most likely scenario is one where all these probabilities are true to some extent. The problem is to identify at which stage of the game the motivational state changes for each player, or which factor dominates others when a player makes a decision.  
	
  Trying to generate a similar trace based on any of these assumptions alone might not give the expected results if the motivation is not identified correctly. For that reason, besides evaluating the current player profile in our experiments, we want to identify possible patterns of behaviour that may have not been yet covered by our current design. A questionnaire such as the one used to collect the player profile (table~\ref{table:playerprofile}) would not be of much help for this purpose, as it would not allow us to identify new factors in players behaviour. A semi-structured interview allows for us to define more general questions to focus the conversation in the player's behaviour, while giving freedom for them to express their ideas.
  
  The script of the interview consists of a set of questions related to the subject's expectations about the game, whether they try to cheat in any way (purposely introduce invalid commands, ask other players how to finish, etc), their opinion on the game they just played, their strategies if they followed any and how they think players with different personalities would have played. Depending on the answers of each player, the interviewer would try to formulate more questions whenever an indicator of a behavior pattern emerged, to get a more detailed explanation of such behavior. The full script of the interview is available as an appendix.

\subsection{Experimental setup}

\begin{table}
\centering
\caption{Statements used to capture the player profile. (n) means that the factor carries a negative score for the corresponding statement.}
\label{table:playerprofile}
\begin{tabularx}{\linewidth}{l X }
\hline
\textbf{Factor}                 & \textbf{Statement}                                                     \\ \hline
f     & My familiarity with the text-based game ``Anchorhead'' is                       \\ \hline
gE                      & My gaming experience is                                                         \\ \hline
gE                      & I think about the consequences of my actions when playing                       \\ \hline
gE (n)            & I complete one quest at a time                                                  \\ \hline
pE           & I explore all the places, elements and characters of the virtual world          \\ \hline
pE           & I complete all quests, including those that aren't necessary to finish the game \\ \hline
pE (n) & I only do what is necessary to pass a level or complete a quest                   \\ \hline
p                     & If I fail a quest, I repeat it until I complete it                              \\ \hline
p                     & I defer my other activities if I'm stuck on a task or mission while playing     \\ \hline
p (n)           & I give up on quests if I find more appealing ones                                \\ \hline
\end{tabularx}
\end{table}

With the author's permission, we made a point-and-click version of Anchorhead based on their code available online \cite{Ontanon}. Using this version, we collected information from 23 anonymous players that generated a total of 36 traces. Before they played, we asked them to fill a questionnaire using the statements in table \ref{table:playerprofile} to calculate their PP. Each of their games was linked to their questionnaire answers using an auto-generated identification number.

The data collected in this experiment is different than the data from our previous work used to identify the PP factors described in our methodology. Three of the participants from the past experiment played the new version of the game. Having recruits re-play the game provided the chance to analyse how players behave when they have previous knowledge of the story.   
After they finished the game(s), players were invited to participate in the interview described previously to obtain more empirical data on their behaviour and validate our approach. Participation in the interview was optional.

Since the experiment could be done online, we do not completely know the players demographics. The target group was anyone over 18 years old and players were invited via social media and printed signs in student areas of our university.
From the 24 total players, 9 agreed to be interviewed; 3 of these participants are female, 6 are male, and their age is between 20 and 35 years. Each interview lasted between 20 and 45 minutes, and the results indicated data saturation, as no new behaviours were identified in the last interviews. 

The questionnaire to calculate PPs consisted of 10 statements where participants indicate their level of agreement from a five-level Likert scale (strongly disagree = 1, strongly agree = 5)~\cite{tullis2010measuring}. To reduce bias in players self-judgement, we introduced statements with a negative score. These statements represent behaviour contrary to what we are looking for in each factor  of the PP, with the exception of familiarity with Anchorhead, as players would remember to have played the game at least once in the past. The statements and the PP factor they measure are shown in table~\ref{table:playerprofile}. The scale of answers for the first two statements goes from `very low' to `very high' rather than `strongly disagree' to `strongly agree'. We abbreviate the factors in the PP as f=familiarity, gE=gaming experience, pE=preference to explore, and p=persistence.

Each PP value collected from the questionnaire is normalised using linear scaling~\cite{aksoy2001feature}. This method obtains a normalised value $N_{x}$ between 0 and 1 applying the following formula: $N_{x}=\frac{x-x_{min}}{x_{max}-x_{min}}$. Substituting $x_{max}=5+5-1=9$ and $x_{min}=1+1-5=3$ as the possible values in our measurements, we normalise each PP factor as $N_{x}=\frac{p^{(1)}_{x}+p^{(2)}_{x}-n^{(1)}_{x}+3}{12}$, where $N_{x}$ is the normalised value for the factor (e.g., persistence), $p^{(1)}_{x}$ and $p^{(2)}_{x}$ are the scores of the two positive statements for that factor in the questionnaire and $n^{(1)}_{x}$ is the score for the negative statement.  

\begin{figure}
\centering
\includegraphics[width=\linewidth]{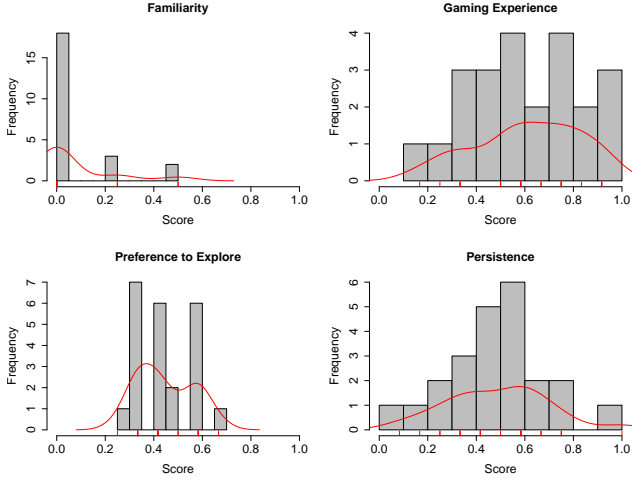}
\caption{Distribution for each factor of the player profiles collected.}
\label{fig:ppDist}
\end{figure}

These normalised values are used in the player profile belief for each player to run simulations with a set of Informed Agents (IAs).
When a player completed the game more than once, we changed their familiarity score to 1 if it was less than 0.5 before their first game, and use the rest of the player profile for their remaining traces. Figure~\ref{fig:ppDist} shows the distribution of scores for each factor of the PPs calculated.
We ran 20 IA simulations using the PP corresponding to each complete player trace. 
We also had 20 simulations of the Uninformed Agent (UA). We compared the similarity of each player trace against the IA and UA traces generated by the simulations.

The elements to compare are the plot points discovered in each trace.
  We consider the Jaccard index the most appropriate way to measure the similarity between traces, as the number of plot points among traces varies. 
The Jaccard index of two sets is defined as the cardinality of their intersection divided by the cardinality of their union, giving a result between 0 and 1~\cite{Romesburg2004}.

\subsection{Findings}
\label{sec:findings}
  
\begin{figure}
\centering
\includegraphics[width=\linewidth]{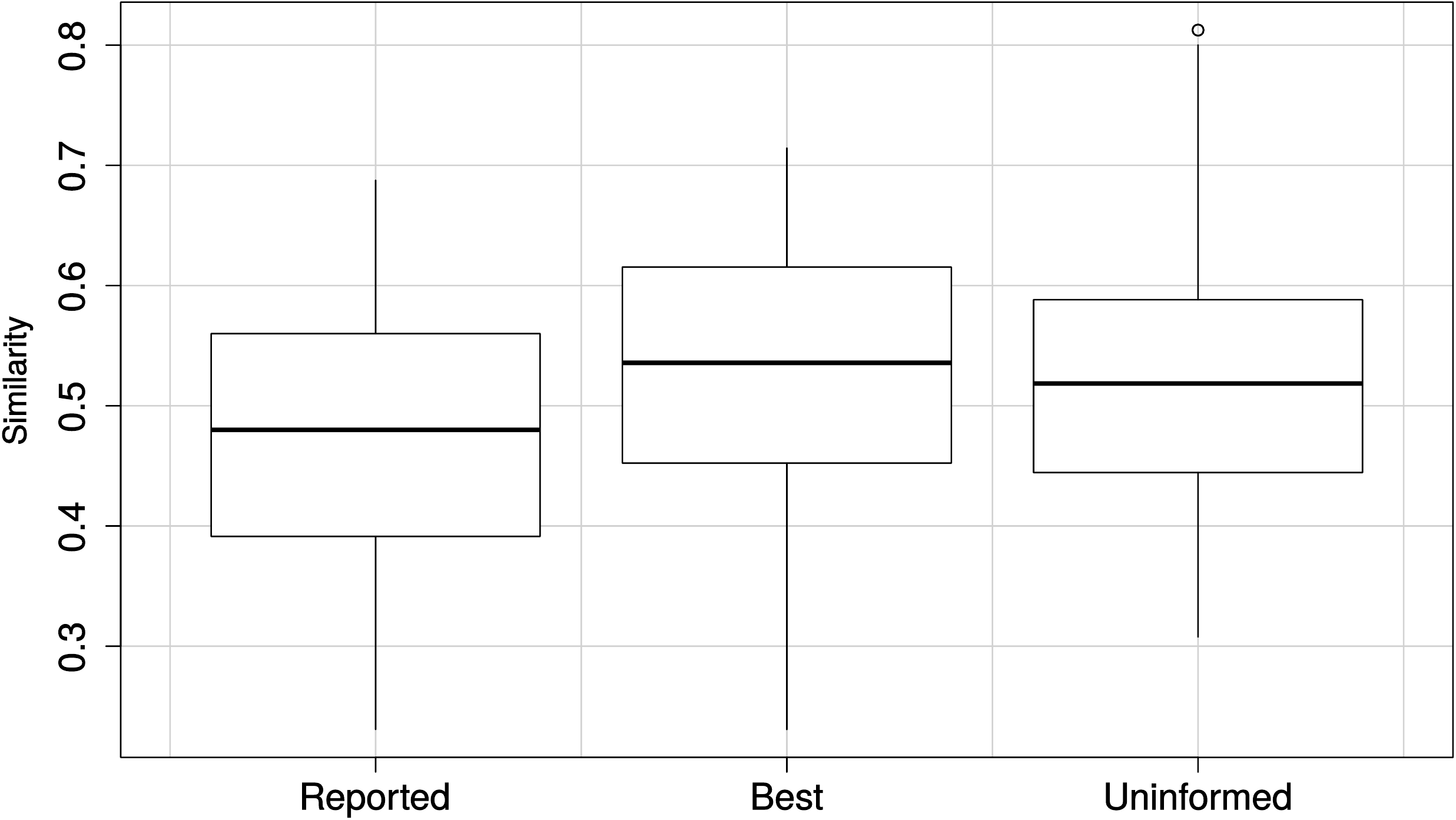}
\caption{General comparison between methods to inform the player model.}
\label{fig:boxplot3Agents}
\end{figure}

\begin{figure*}
\centering
\includegraphics[width=\linewidth]{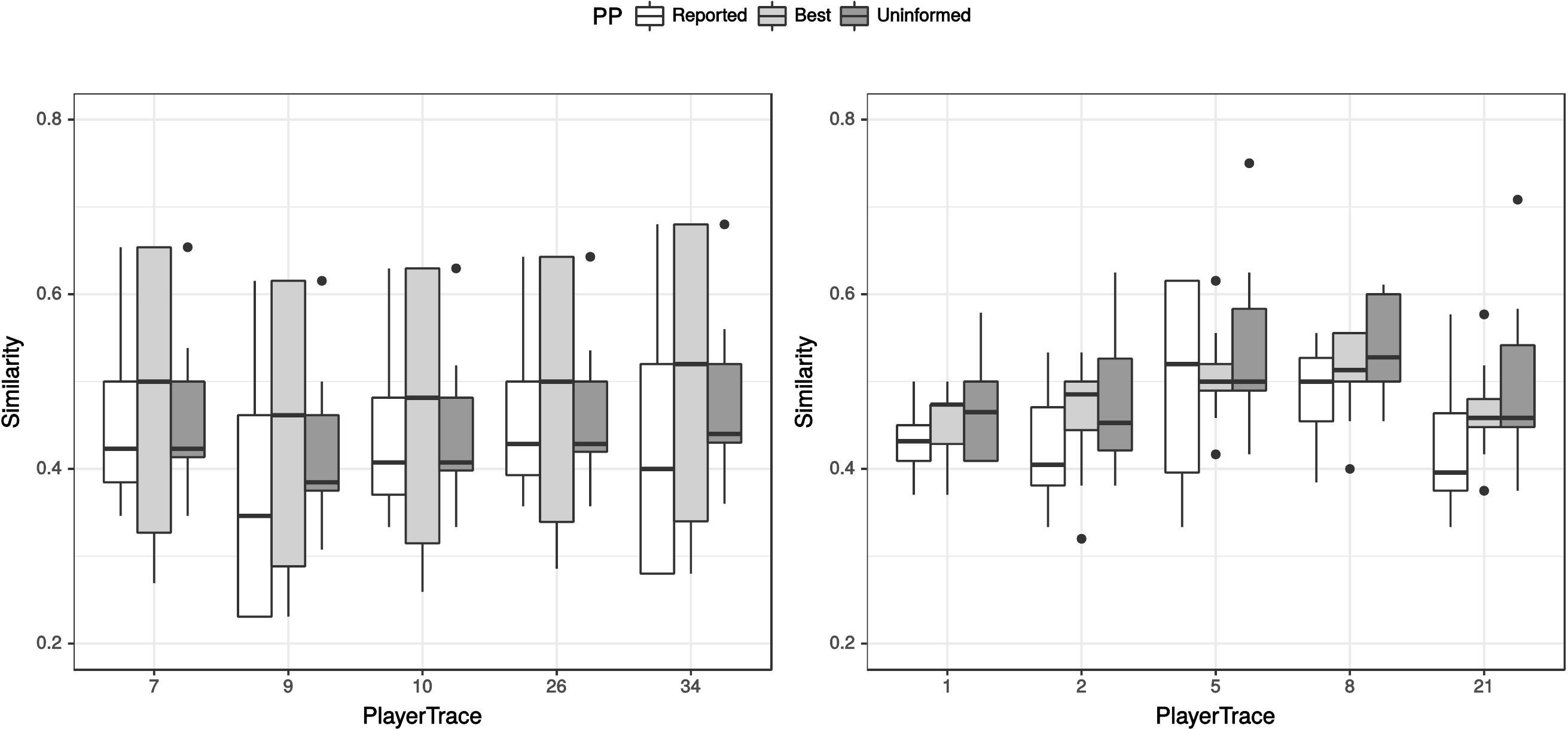}
\caption{Highest (left) and lowest (right) performance of the IA using the best PP vs UA.}
\label{fig:boxPlotCompare}
\end{figure*}

% \usepackage{graphicx}
% \usepackage{booktabs}

%\begin{table}
%\centering
%\caption{Similarities of traces collected with IA and UA traces. Only reporting instances where the PP obtained from players answers matches the best PP found with our simulations.}
%\label{table:summarisedSimComparison}
%\resizebox{\columnwidth}{!}{%
%\begin{tabular}{p{0.113\columnwidth}p{0.035\columnwidth}p{0.054\columnwidth}p{0.054\columnwidth}p{0.035\columnwidth}p{0.104\columnwidth}p{0.104\columnwidth}p{0.104\columnwidth}p{0.104\columnwidth}p{0.104\columnwidth}p{0.104\columnwidth}} 
%\toprule
%                   & \multicolumn{7}{p{0.489\columnwidth}}{Informed} & \multicolumn{3}{p{0.312\columnwidth}}{Uninformed}  \\ 
%\midrule
%Player\par{} trace & f & gE & pE & p & Min   & Mean  & Max           & Min   & Mean  & Max                                \\ 
%\midrule
%19                 & 1 & 1  & 0  & 1 & 0.24  & 0.435 & 0.577         & 0.32  & 0.436 & 0.64                     %          \\
%24                 & 0 & 1  & 0  & 1 & 0.545 & 0.621 & 0.652         & 0.458 & 0.573 & 0.652                    %          \\
%28                 & 1 & 1  & 0  & 1 & 0.231 & 0.419 & 0.556         & 0.308 & 0.419 & 0.615                    %          \\
%31                 & 1 & 1  & 0  & 1 & 0.346 & 0.533 & 0.667         & 0.423 & 0.535 & 0.731                              \\
%\bottomrule
%\end{tabular}
%}
%\end{table}

% \usepackage{multirow}
% \usepackage{booktabs}

% \usepackage{booktabs}

\begin{table}
\centering
\caption{Similarities of traces collected with IA and UA traces. Only reporting instances where the PP obtained from players answers matches the best PP found with our simulations.}
\label{table:summarisedSimComparison}
\resizebox{\columnwidth}{!}{%
\begin{tabular}{lllllllllll} 
\toprule
Player & \multicolumn{4}{l}{Player profile } & \multicolumn{3}{l}{Informed} & \multicolumn{3}{l}{Uninformed}  \\ 
\cline{2-11}
trace  & f & gE & pE & p                     & Min   & Mean  & Max          & Min   & Mean  & Max             \\ 
\midrule
19     & 1 & 1  & 0  & 1                     & 0.24  & 0.435 & 0.577        & 0.32  & 0.436 & 0.64            \\
24     & 0 & 1  & 0  & 1                     & 0.545 & 0.621 & 0.652        & 0.458 & 0.573 & 0.652           \\
8      & 1 & 1  & 0  & 1                     & 0.231 & 0.419 & 0.556        & 0.308 & 0.419 & 0.615           \\
1      & 1 & 1  & 0  & 1                     & 0.346 & 0.533 & 0.667        & 0.423 & 0.535 & 0.731           \\
\bottomrule
\end{tabular}
} %end of resize box
\end{table}

The results of the initial observations of IAs and UAs are reported in figures \ref{fig:boxplot3Agents} and \ref{fig:boxPlotCompare}, with the box plots marked as ``Reported'' and ``Uninformed''.

Since this comparison did not indicate better results using the PP in the IAs, we analysed some player traces to confirm if their actions matched their answers to the questionnaire. We found that players who were behaving like explorers, for instance, did not give answers that would result in a high score for the corresponding element of the PP. We found similar cases for the other PP factors.
We cross-checked this issue with information from the interviews. Some players mentioned that even though their goal was not to find every item available, they constantly checked if there is something they may need in the future, so they tried to exhaust the options available in their current location before moving to a different area.

To find out if a more accurate PP would deliver a more similar trace, we ran simulations for all the 16 possible combinations of PPs. Since the IA's model only differentiates values $N_x \leq 0.5$ or $N_x>0.5$, we generated  20 traces with binary values for each $PP = \{f, gE, pE, p\}$ between \{0,0,0,0\}, \{0,0,0,1\}, ... \{1,1,1,1\}.
To find the best matching PP, we selected the set of traces with the highest mean of similarities between this new set of IAs and the real player traces. The results are shown with the box plots marked as ``Best'' in figures \ref{fig:boxplot3Agents} and \ref{fig:boxPlotCompare}. 

Figure \ref{fig:boxplot3Agents} summarises the results for all the player traces we used to compare our methods.
While the UAs produced higher maximum values, the first and third quartiles are slightly lower than those produced with the best matching PP. This means that that half of the data is concentrated in a better range for the agents informed with the best PP. Figure \ref{fig:boxplot3Agents} also shows the lower performance when using the PP reported by players in their questionnaire.
Figure \ref{fig:boxPlotCompare} gives a more detailed comparison of each agent's performance in a subset of cases. We selected the observations with the five highest and lowest performance of the IA using the best PP over the UA. In this figure, it is easier to appreciate that the maximum values obtained with the UA tend to be outliers of a smaller distribution with lower similarities than the IA. At the same time, the IA's larger distribution size shows that it is unable to reproduce the players behaviour consistently when reaching higher similarities.

In contrast, the lowest performing IAs seem to produce a more consistent behaviour across simulations, as their distribution size is much smaller than the highest performance ones. A positive observation is that in most of these cases, the first quartile and median are at least the same as for the UAs.
We will now discuss other findings that help us explain why the results obtained with the IAs are not much better than with UAs.

As pointed out previously, the players' answers to the questionnaire were not accurate enough to produce a representative player profile in most cases. Table~\ref{table:summarisedSimComparison} shows the only four instances (out of 36) where the PP calculated from players responses is the same as the best matching PP identified with our simulations. As BDI highly relies on the expert's perception of the subjects decision making, the correct identification of the type of subject being simulated is crucial to produce good results.

%Inexperienced players are less reliable (Similarities with low gExp in reported pp).
In all the cases where the reported PP was the best possible, the players gaming experience is set to 1, which means the calculated score from their answers was $\geq 0.5$. This could be an indicator that experienced players know their behaviour better than novices. Another interesting fact is that all these cases share the same values for preference to explore and persistence besides gaming experience. The fact that only one of these four players was not familiar with the game can also mean that these players knew how to answer for this specific game because they have already observed themselves by playing this or similar IN games before.

%Order of element exploration affects plotpoints discovered (Players interviewed say their behaviour would have been different if they found an element or location 
Players interviewed described their strategies, and the player model implements most of them depending on the PP. There is a behaviour pattern reported by all of the players interviewed, and observed in the traces collected. At the start of the game, players are located in the livingroom and have the option to go either to the hall of the mansion or to the street. In their first game, all the players chose to move to the hall rather than going outside. Information obtained from the interviews suggests that this behaviour is a result of players applying the concept of being indoors within a virtual world, even when factors such as distance or weather do not affect them (e.g. they do not lose stamina by walking, or life points if it is raining).
 In our case study, this means that most real players will discover clues and plot points from inside the building that will trigger goals related to those items first. On the other hand, our player model gives priority to places that haven't been explored, but when all the places available have the same status, this choice is made at random. If the agents go to the street first, the trace generated will be different regardless of the accuracy of the PP, as the goals triggered early in the game were different in the first place.

%Pleople who played more than twice (usually high prefEx and/or exp) cannot be modelled by this current work.
The player model only considers re-playing the game once, as there are some plans that run differently depending on the value of the familiarity factor. However, some of our recruits played more than two times. The game had two possible endings, but recruits were not informed about this before they played. Those who played more than twice reported to do so because they had objects whose purpose they had not found yet. This behavior was reported in players with high experience with INs; they mentioned that according to their experience, every item usually has a purpose, and some of them usually continue playing until they find it. Our game had a few items that did not contribute to unlocking an ending or solving a mystery, some of them only had the purpose of letting the player know whose character they belonged to, but after that the object could not be traded or used anywhere else. 

%Player goals change during the game.
There were cases where players reported to completely change their mind about which goal to pursue, e.g., they are trying to find out who William is and they have the goal to ask characters about him, then they discover the observatory and now pursue the goal of finding the lens for the telescope. While the persistence factor was considered with this behaviour in mind, we do not have the information to decide exactly when to drop one goal, or which one to pursue with higher priority than the rest. Our current model only drops a goal if it has been attempted for long enough when the persistence score is low, but prioritising the current goals specifically for each subject is a challenging problem.

%Having played the game only once before affects current behaviour more than currently modelled. 
We also observed that players who knew the story (even from long ago and claimed not to remember the details) behaved substantially differently than those completely new to the game. Some players who reported low familiarity with Anchorhead because they argued to have lost memory of it generated traces with less erratic behaviour. Players who were interviewed reported that the dynamics slowly came back to them as they played. Moreover, the best PP found for their first trace has a value of 1 in familiarity. This means that this factor should not be calculated with a Likert scale, but as a boolean.

\balance{}
\section{Conclusion and Future work}
\label{sec:conclusion}
We have presented a Belief-Desire-Intention (BDI) player model design for Interactive Narratives (INs) that makes use of a Player Profile (PP) as part of its belief set to select from different plans available to achieve the same goal. Our aim was to create game traces as similar as possible to the trace generated by the player whose PP was being used.
We explained our PP and performed a mixed analysis combining empirical evidence from player traces, semi-structured interviews and the measured similarity between real player traces and simulated traces.

The information learned from the interviews with our players seems to confirm that the factors on the player profile are useful, as they define the behaviour of our recruits within the game.
However, our analysis implementation of the profile in the BDI player model doesn't seem to deliver better results than using an uninformed BDI model, where no information from the player is known and there is only one way to achieve a goal. We explained in our analysis our thoughts about why we were not successful in replicating players behaviour with more accuracy despite having more information about them.

Regarding the first research aim in the introduction of this paper, our evidence suggests that we were able to identify the factors that define players behaviour in an IN. However, players were not a reliable source to report their own behavioural patterns, making the questionnaire used a sub-optimal tool to calculate the PP. As for our second research aim, besides being able to properly measure the PP factors, we found that there are strategies and factors that need to be taken into account before we can find a definitive answer. We emphasise that simulating the behaviour of specific players is a very complex task, as their motives and intentions are dynamic most of the time despite having a reasonable understanding about their preferences.

Our contributions are the validation of our PP via empirical evidence, a methodology for designing and implementing a BDI player model of an IN, and the observations listed in the findings section. As future work, we aim to develop a new version of the informed player model, where we implement the information learned from this work's analysis. For example, a memory of plot points and elements discovered in previous games in order to mimic goals such as exhausting the options available in the virtual world. A better approach to prioritise and/or change goals being pursued is also needed.

A possible solution to the players reliability problem would be to predict new players traces without the need of data collection prior to the game. We were able to identify similar behaviours among players that shared scores for some of the profile factors. With the right approach to measure the scores for the player profile, Plan Recognition (PR) could be a way to create and update a player profile using data collected from other players in the past.
Finally, we would also like to investigate the efficiency of apprenticeship learning to aid the PR task. Learning from demonstration seems to be a good way to reduce the subjectivity introduced by BDI modellers, hence learning behaviour common among real players in situations not considered during the model design. This can help solve problems like discovering the wrong plot points early in the game due to random choices that humans would not make.

%\begin{comment}
\balance{}
\section{Acknowledgements}
J. Rivera-Villicana is financially supported by the National Council of Science and Technology in Mexico (CONACyT).
%\end{comment}

% Balancing columns in a ref list is a bit of a pain because you
% either use a hack like flushend or balance, or manually insert
% a column break.  http://www.tex.ac.uk/cgi-bin/texfaq2html?label=balance
% multicols doesn't work because we're already in two-column mode,
% and flushend isn't awesome, so I choose balance.  See this
% for more info: http://cs.brown.edu/system/software/latex/doc/balance.pdf
%
% Note that in a perfect world balance wants to be in the first
% column of the last page.
%
% If balance doesn't work for you, you can remove that and
% hard-code a column break into the bbl file right before you
% submit:
%
% http://stackoverflow.com/questions/2149854/how-to-manually-equalize-columns-
% in-an-ieee-paper-if-using-bibtex
%
% Or, just remove \balance and give up on balancing the last page.
%
%\balance{}

% BALANCE COLUMNS
\balance{}

% REFERENCES FORMAT
% References must be the same font size as other body text.
\bibliographystyle{sigchi-latex-proceedings/SIGCHI-Reference-Format}
\bibliography{library}

\end{document}

\title{Script for semi-structured interview after recruits played the game}
\author{}
\date{}
\maketitle
\thispagestyle{empty}

\begin{enumerate}
\item Quick introduction between PhD student and recruit.
\item What is the player ID the game generated for you?
\item What kinds of things were you expecting from the game, in terms of:
  \begin{enumerate}
  \item Game play (Whether the game was about treasure-hunting, puzzles, chasing, etc.).
  \item Objectives of the game (Points, win/lose, competition with others).
  \end{enumerate}
\item Did you try to break the game? (e.g. Introduce invalid commands on purpose)
  \begin{enumerate}
  \item What sort of holes did you find?
  \item Were you successful in breaking the game?
  \end{enumerate}  
\item Tell me about your experience of playing the game.
  \begin{enumerate}
  \item What worked/didn't work?
  \item What did you like/didn't like?
  \item How was it different to your expectation?
  \item Do you think the actions changed the direction of the story line?
  \item Would/did you go back and play it differently to see different parts of the story?
  \item Were you interested in the story to see what happened next/at the end?  
  \end{enumerate}
\item After playing, would you describe yourself as :
  \begin{enumerate}
  \item An experienced gamer?
  \item Typical gamer (How do you see typical players)?
  \end{enumerate}
  
\item What do you do when you play a game for the first time? (In terms of strategy)
\item What do you do when you fail a mission or quest?
  \begin{enumerate}
  \item If you can’t advance through the game without that mission.
  \item If the mission is optional.
  \end {enumerate}
\end{enumerate}